\documentclass[submission,copyright,creativecommons]{eptcs}
\providecommand{\event}{TARK 2015} 
\usepackage{graphicx}
\usepackage{pstricks}
\usepackage{color}
\usepackage[all]{xy}

\usepackage{auto-pst-pdf}

\usepackage{amsmath}
\usepackage{amsfonts}
\usepackage{amssymb,bm}
\usepackage{verbatim}
\usepackage{stmaryrd}
\usepackage{latexsym}
\usepackage{times}
\usepackage{enumerate}

\usepackage{graphicx}     

\newcommand{\NI}{\noindent}
\newcommand{\bfe}[1]{\begin{bfseries}\emph{#1}\end{bfseries}\index{#1}}
\newcommand{\LL}{\mbox{$\ldots$}}
\newcommand{\sse}{\mbox{$\:\subseteq\:$}}
\newcommand{\HB}{\hfill{$\Box$}}
\newcommand{\C}[1]{\mbox{$\{{#1}\}$}}   
\newcommand{\Ra}{\mbox{$\:\Rightarrow\:$}}

\newcommand{\Atomsb}{\mathsf{Q}}
\newcommand{\Agents}{\mathsf{A}}
\newcommand{\Atoms}{\mathsf{P}}

\newcommand{\newproof}[3]{
   \newenvironment{#1}[1][]%
   {%
      \begin{trivlist}%
         \item[\hspace{\labelsep}\textnormal{\textbf{#2%
            \def\op@@@arg{##1}%
            \ifx\op@@@arg\empty
            \else~(##1)\fi
         }}]%
   }%
   {%
      #3
      \end{trivlist}%
   }%
}
\newproof{proof}{Proof}{\qed}
\def\qed{{\ifhmode\unskip\nobreak\hfil\penalty50 \hskip1em \else\nobreak\fi
   \mbox{}\nobreak\hfil\qedtext%
   \parfillskip=0pt \finalhyphendemerits=0 \par}}

\def\qedtext{\ensuremath{\square}}
\RequirePackage{amsfonts}
\DeclareMathSymbol{\square}       {\mathord}{AMSa}{"03}

\newtheorem{theorem}{Theorem}[section]
\newtheorem{definition}[theorem]{Definition}
\newtheorem{example}[theorem]{Example}
\newtheorem{fact}[theorem]{Fact}
\newtheorem{proposition}[theorem]{Proposition}

\newcommand{\pull}{\triangleleft}
\newcommand{\push}{\triangleright}

\newcommand{\set}[1]{{\{ #1 \}}}

\newcommand{\Model}{\mathcal{M}}
\newcommand{\Situations}{\mathsf{S}}

\newcommand{\Situation}{\mathsf{s}}

\newcommand{\init}{\mathsf{root}}
\newcommand{\Call}{\mathsf{c}}
\newcommand{\Calld}{\mathsf{d}}

\newcommand{\CSequences}{{\bf C}}

\newcommand{\CSequence}{{\bf c}}
\newcommand{\CSequenced}{{\bf d}}

\title{Epistemic Protocols for Distributed Gossiping}
\author{Krzysztof R. Apt
\institute{Centrum Wiskunde Informatica}
       \institute{Amsterdam, The Netherlands}
       \email{k.r.apt@cwi.nl}
\and
Davide Grossi \qquad\qquad Wiebe van der Hoek
\institute{University of Liverpool} 
\institute{Liverpool, UK}
\email{\quad d.grossi@liv.ac.uk \quad\qquad wiebe@liv.ac.uk}
}
\def\titlerunning{Epistemic Protocols for Distributed Gossip}
\def\authorrunning{K. Apt, D. Grossi \& W. van der Hoek}

\begin{document}
\maketitle

\begin{abstract}
Gossip protocols aim at arriving, by means of point-to-point or group communications, at a situation in which all the agents know each other's secrets. We consider distributed gossip protocols which are expressed by means of epistemic logic.  We provide an operational semantics of such protocols and set up an appropriate framework to argue about their correctness. Then we analyze specific protocols for complete graphs and for directed rings.
\end{abstract}

\section{Introduction}
In the gossip problem (\cite{tijdeman:1971,baker72gossips}, see also
\cite{HHL88} for an overview) a number $n$ of agents, each one
knowing a piece of information (a \emph{secret}) unknown to the
others, communicate by one-to-one interactions (e.g., telephone calls).
The result of each call is that the two agents involved in it learn
all secrets the other agent knows at the time of the call. The problem
consists in finding a sequence of calls which disseminates all the
secrets among the agents in the group.  It sparked a large literature
in the 70s and 80s
\cite{tijdeman:1971,baker72gossips,hajnal72cure,bumby81problem,seress86quick}
typically focusing on establishing---in the above and other variants
of the problem---the minimum number of calls to achieve dissemination
of all the secrets. This number has been proven to be $2n - 4$, where
$n$, the number of agents, is at least 4.

The above literature assumes a centralized perspective on the gossip
problem: a planner schedules agents' calls. In this paper we pursue a
line of research first put forth in \cite{ADGH14} by developing a
decentralized theory of the gossip problem, where agents perform calls
not according to a centralized schedule, but following individual epistemic
protocols they run in a distributed fashion. These protocols tell the
agents which calls to execute depending on what they know, or do not
know, about the information state of the agents in the group. We
call the resulting distributed programs \emph{(epistemic) gossip protocols}.

\paragraph{Contribution of the paper and outline}
The paper introduces a formal framework for specifying epistemic gossip
protocols and for studying their computations in terms of correctness,
termination, and fair termination (Section
\ref{section:preliminaries}). It then defines and studies two natural
protocols in which the interactions are unconstrained (Section
\ref{section:complete}) and four example gossip protocols in which
agents are positioned on a directed ring and calls can happen only
between neighbours (Section \ref{section:rings}).  Proofs are
collected in the appendix.

From a methodological point of view, the paper integrates concepts and
techniques from the distributed computing, see, e.g., \cite[Chapter
11]{ABO09} and the epistemic logic literature
\cite{fagin95reasoning,meyer95epistemic} in the tradition of
\cite{parikhetal:1985,kurki-suonio86towards,fagin97knowledge}.

\section{Gossip protocols} \label{section:preliminaries}

We introduce first the syntax and semantics of gossip protocols.

\subsection{Syntax}

We loosely use the syntax of the language CSP (Communicating
Sequential Processes) of \cite{Hoa78} that extends the guarded
command language of \cite{Dij75} by disjoint parallel composition and
commands for synchronous communication.  CSP was realized in
the distributed programming language OCCAM (see INMOS \cite{INM84}).

The main difference is that we use as guards epistemic formulas and as
communication primitives calls that do not require synchronization.
Also, the syntax of our distributed programs is very limited.  
In order to define gossip protocols we introduce in turn calls and epistemic guards.  

\smallskip

Throughout the paper we assume a fixed finite set $\mathsf{A}$ of at
least three \bfe{agents}. We assume that each agent holds exactly one
\bfe{secret} and that there exists a bijection between the set of
agents and the set of secrets.  We denote by $\mathsf{P}$ the set of
all secrets (for \emph{p}ropositions).  Furthermore, it is assumed
that each secret carries information identifying the agent to whom
that secret belongs.


\subsubsection{Calls} 

Each \bfe{call} concerns two agents, the \emph{caller} ($a$ below) 
 and the \emph{agent called} ($b$).
We distinguish three \bfe{modes of communication} of a call:

\begin{description}
\item{\bfe{push-pull}}, written as $ab$ or $(a,b)$. 
During this call the caller and the called agent learn each other's secrets,
\item{\bfe{push}}, written as $a \push b$. 
After this call the called agent learns all the secrets held by the
  caller,
\item{\bfe{pull}}, written as $a \pull b$. 
After this call the caller learns all the secrets held by the called
agent.
\end{description}
Variables for calls are denoted by $\Call$, $\Calld$.  Abusing
notation we write $a \in \Call$ to denote that agent $a$ is one of the
two agents involved in the call $\Call$ (e.g., for $\Call := ab$ we
have $a \in \Call$ and $b \in \Call$).  
Calls in which agent $a$ is involved are denoted by $\Call^a$.

\subsubsection{Epistemic guards}
 
Epistemic guards are defined as formulas in a simple modal language with the following grammar:
\[
\phi ::= F_a p \mid \neg \phi \mid \phi \land \phi \mid K_a \phi,
\]
where $p \in \mathsf{P}$ and $a \in \mathsf{A}$.  Each secret is
viewed as a distinct symbol.
We denote the secret of agent $a$ by $A$, the secret of agent $b$ by $B$ and so on. 
We denote the set of so defined formulas by ${\mathcal L}$ and we refer to
its members as epistemic formulas or epistemic guards.  We read $F_a
p$ as `agent $a$ is familiar with the secret $p$' 
(or `$p$ belongs to the set of secrets $a$ knows about')
and $K_a \phi$ as `agent $a$
knows that formula $\phi$ is true'. So this language is an epistemic
language where atoms consist of `knowing whether' statements about
propositional atoms, if we view secrets as Boolean variables.

Atomic expressions in $\mathcal{L}$ concern only who knows what
secrets. As a consequence the language cannot express formally the
truth of a secret $p$. This level of abstraction suffices for the
purposes of the current paper.  However, expressions $F_a p$ could be
given a more explicit epistemic reading in terms of `knowing whether'.
That is, `$a$ is familiar with $p$' can be interpreted (on a suitable
Kripke model) as `$a$ knows whether the secret $p$ is true or not'.
This link is established in \cite{ADGH14}.

\subsubsection{Gossip protocols}

Before specifying what a program for agent $a$ is, let us first define
the language ${\mathcal L}_a$ with the following grammar:
\[
\psi ::= K_a \phi \mid \neg \psi \mid \psi \land \psi
\]
with $\phi \in {\mathcal L}$.\footnote{Alternatively, $\mathcal{L}_a$
  could be defined as the fragment of $\mathcal{L}$ consisting of the
  formulae of form $K_a \psi$. In logic S5, it is easy to prove that
  each $\psi \in \mathcal{L}_a$ is logically equivalent to a formula
  $K_a \phi \in \mathcal{L}$.}

By a \bfe{component program}, in short a \bfe{program}, for an agent $a$ we mean
a statement of the form
\[ 
*[[]^m_{j=1}\ \psi_j \to \Call_j],
\]
where $m > 0$ and each $\psi_j \to \Call_j$ is such that $\psi_j \in
{\mathcal L}_a$ and $a$ is the caller in $\Call_j$.

Given an epistemic formula $\psi \in {\mathcal L}_a$ and a call $\Call$, we call the
construct $\psi \to \Call$ a \bfe{rule} and refer in this context to
$\psi$ as a \bfe{guard}.  

We denote the set of rules $\{\psi_1 \to \Call_1, \LL, \psi_k \to \Call_k\}$
as
$[[]^k_{j=1}\ \psi_j \to \Call_j]$ and abbreviate a set of rules
$\{\psi_1 \to \Call, \LL, \psi_k \to \Call\}$ with the same call to a single
rule $\bigvee_{i = 1}^{k} \psi_i \to \Call$.


Intuitively, $*$ denotes a repeated execution of the rules, one at a time, where
each time a rule is selected whose guard is true.

Finally, by a \bfe{distributed epistemic gossip protocol}, in short a
\bfe{gossip protocol}, we mean a parallel composition of component
programs, one for each agent.  In order not to complicate matters we
assume that each gossip protocol uses only one mode of communication.

Of special interest for this paper are gossip protocols that are symmetric. By
this we mean that the protocol is a composition of the component
programs that are identical modulo the names of the agents. Formally,
consider a statement $\pi(x)$, where $x$ is a variable ranging over
the set $\mathsf{A}$ of agents and such that for each agent $a \in
\mathsf{A}$, $\pi(a)$ is a component program for agent $a$.  Then the
parallel composition of the $\pi(a)$ programs, where $a \in
\mathsf{A}$, is called a \bfe{symmetric gossip protocol}.

Gossip protocols are syntactically extremely simple. Therefore it
would seem that little can be expressed using them.  However, this is
not the case. In Sections \ref{sec:complete} and \ref{sec:ring} we
consider gossip protocols that can exhibit complex behaviour.

\subsection{Semantics}

We now move on to provide a formal semantics of epistemic guards, and
then describe the computations of gossip protocols.

\subsubsection{Gossip situations and calls}

A \bfe{gossip situation} is a sequence $\Situation = (\Atomsb_a)_{a
  \in \Agents}$, where $\Atomsb_a \sse \Atoms$ for each agent $a$.
Intuitively, $\Atomsb_a$ is the set of secrets $a$ is familiar with in situation $\Situation$.  
  The \bfe{initial gossip situation} is the one in which each
$\Atomsb_a$ equals ${\{A\}}$ and is denoted by $\init$. The set of all
gossip situations is denoted by $\Situations$. We say that an agent $a$
is an \bfe{expert} in a gossip situation $\Situation$ if he is familiar in
$\Situation$ with all the secrets, i.e., if $\Atomsb_a = \Atoms$.
The initial gossip situation reflects the fact that initially each
agent is familiar only with his own secret, although it is not assumed this is
common knowledge among the agents. In fact, in the introduced
language we have no means to express the concept of common knowledge. 

\smallskip

We will use the following concise notation for gossip situations. Sets
of secrets will be written down as lists. e.g., the set $\set{A, B,
  C}$ will be written as $ABC$. Gossip situations will be written down as lists of
lists of secrets separated by dots. E.g., if there are three
agents, $\init = A.B.C$ and the situation $(\set{A,B},
\set{A,B}, \set{C}$) will be written as $AB.AB.C$.


Each call transforms the current gossip situation by modifying the set
of secrets the agents involved in the call are familiar with. More
precisely, the application of a call to a situation is defined as follows.

\begin{definition}[Effects of calls] \label{def:effects}
A call is a function $\Call: \Situations \longrightarrow \Situations$, so defined,  for $\Situation := (\Atomsb_a)_{a \in \Agents}$:
\begin{description}

\item[\fbox{$\Call = ab$}] $\Call(\Situation) = (Q'_a)_{a \in \Agents}$, where 
$\Atomsb'_a = \Atomsb'_b = \Atomsb_a \cup \Atomsb_b$,
$\Atomsb'_c = \Atomsb_c$, for $c \neq a,b$;

\item[\fbox{$\Call = a \push b$}] $ \Call(\Situation) =(Q'_a)_{a \in \Agents}$, where
$\Atomsb'_b = \Atomsb_a \cup \Atomsb_b$,
$\Atomsb'_a = \Atomsb_a$,
$\Atomsb'_c = \Atomsb_c$, for $c \neq a,b$;

\item[\fbox{$\Call = a \pull b$}] $\Call(\Situation) = (Q'_a)_{a \in \Agents}$, where
$\Atomsb'_a = \Atomsb_a \cup \Atomsb_b$,
$\Atomsb'_b = \Atomsb_b$,
$\Atomsb'_c = \Atomsb_c$, for $c \neq a,b$.
\end{description}
\end{definition}
The definition formalizes the modes of communications we introduced earlier.
Depending on the mode, secrets are either shared between
caller and callee ($ab$), they are pushed from the caller to the
callee ($a \push b$), or they are retrieved by the caller from the
callee ($a \pull b$). 

\subsubsection{Call sequences}

A \bfe{call sequence} is a (possibly infinite) sequence of calls, in symbols $(\Call_1, \Call_2, \ldots, \Call_n, \ldots)$, all
being of the same communication mode. 
The empty sequence is denoted by
$\epsilon$.  We use $\CSequence$ to denote a call sequence and
$\CSequences$ to denote the set of all call sequences. The set of all
finite call sequences is denoted $\CSequences^{< \omega}$.  Given a
finite call sequence $\CSequence$ and a call $\Call$ we denote by
$\Call.\CSequence$ the prepending of $\CSequence$ with $\Call$,
and  by $\CSequence.\Call$ the postpending of $\CSequence$
with $\Call$.

The result of applying a call sequence to a situation $\Situation$ is
defined by induction using Definition \ref{def:effects}, as follows:

\NI
[Base] $\epsilon(\Situation) := \Situation$,

\NI
[Step] $(\Call.\CSequence)(\Situation) := \CSequence(\Call(\Situation))$.

\begin{example}
\rm

Let the set of agents be $\set{a,b,c}$.
\[
\arraycolsep=1.3pt\def\arraystretch{1.5}
\begin{array}{ccccccc}   
          &ab&                        &ca&                               &ab& \\
A.B.C&             &AB.AB.C&            &ABC.AB.ABC&           &ABC.ABC.ABC
\end{array}
\]
The top row lists the call sequence $(ab,ca,ab)$, while
the bottom row lists the successive gossip situations obtained from the initial
situation $A.B.C$ by applying the calls in the sequence: first $ab$, then $ca$ and finally
$ab$.
\HB
\end{example}

By applying an infinite call sequence $\CSequence = (\Call_1, \Call_2,
\ldots, \Call_n, \ldots)$ to a gossip situation $\Situation$ one
obtains therefore an infinite sequence $\CSequence^0(\Situation),
\CSequence^1(\Situation), \ldots, \CSequence^n(\Situation), \ldots$ of
gossip situations, where each $\CSequence^k$ is sequence $\Call_1, \Call_2, \ldots,
\Call_k$. A call sequence $\CSequence$ is said to \bfe{converge} if for
all input gossip situations $\Situation$ the generated sequence of gossip
situations reaches a limit, that is, there exists $n < \omega$ such
that for all $m \geq n$ $\CSequence^m(\Situation) =
\CSequence^{m+1}(\Situation)$. 
Since the set of secrets is finite and calls never make agents forget
secrets they are familiar with, it is easy to see the following.
\begin{fact}
All infinite call sequences converge.
\end{fact}

However, as we shall see, this does not imply that all gossip protocols terminate.
In the remainder of the paper, unless stated otherwise, we will assume the push-pull mode of communication.
The reader can easily adapt our presentation to the other modes.


\subsubsection{Gossip models}

The set $\Situations$ of all gossip situations is the set of all possible combinations of secret distributions among the agents.
As calls progress in sequence from the initial situation, agents may
be uncertain about which one of such secrets distributions is the
actual one. This uncertainty is precisely the object of the epistemic
language for guards we introduced earlier.



\begin{definition} \label{def:model}
A \bfe{gossip model} (for a given set $\Agents$) is a tuple $\Model = (\CSequences^{< \omega}, \set{\sim_a}_{a \in \Agents})$, where each $\sim_a \subseteq
\CSequences^{< \omega} \times \CSequences^{< \omega}$ is the smallest relation satisfying the following inductive conditions
(assume the mode of communication is push-pull):
\begin{description}
\item{{\em [Base]}} $\epsilon \sim_a \epsilon$; 
\item{{\em [Step]}} Suppose $\CSequence \sim_a \CSequenced$.
 \begin{enumerate}[(i)]
  \item If $a \not\in \Call$, then $\CSequence.\Call \sim_a \CSequenced$
and $\CSequence \sim_a \CSequenced.\Call$.
\item If there exists $b \in \Agents$ and $\Call,\Calld \in \set{ab, ba}$ such that  $\CSequence.\Call(\init)_a = \CSequenced.\Calld(\init)_a$, then $\CSequence.\Call \sim_a \CSequenced.\Calld$.
  \end{enumerate}
\end{description}
A gossip model with a designated finite call sequence is called a
\bfe{pointed gossip model}.

For the push, respectively pull, modes of communication clause (ii) needs to be
modified by requiring that for some $b \in \Agents$, $\Call = \Calld =
a \push b$ or $\Call = \Calld = a \pull b$, respectively.  
\end{definition}

For instance, by \emph{(i) }we have $ab, bc \sim_a ab, bd$.
But we do not have $bc, ab \sim_a bd, ab$ since
$(bc, ab)(\init)_a$ $= ABC \neq ABD = (bd, ab)(\init)_a$.

Let us flesh out the intuitions behind the above definition. 
Gossip models are needed in order to interpret the epistemic guards of gossip protocols. Since such guards are relevant only after finite sequences of calls, the domain of a gossip model is taken to consist only of finite sequences. Intuitively, those are the finite sequences that can be generated by a gossip protocol.
Let us turn now to the $\sim_a$ relation. This is defined with the following intuitions in mind.
First of all, no
agent can distinguish the empty call sequence from itself---this is
the base of the induction.  Next, if two call sequences are
indistinguishable for $a$, then the same is the case if \emph{(i)} we
extend one of these sequences by a call in which $a$ is not involved
or if \emph{(ii)} we extend each of these sequences by a call of $a$
with the same agent (agent $a$ may be the caller or the callee),
provided $a$ is familiar with exactly the same secrets after each of
the new sequences has taken place---this is the induction
step.\footnote{Notice that the definition requires a designated
 initial situation, which we assume to be $\init$.} 
  

The above intuitions are based on the following
assumptions on the form of communication we presuppose: (i) At the
initial situation, as communication starts, each agent knows only her
own secret but considers it possible that the others may be familiar
with all other secrets. In other words there is no such thing as
common knowledge of the fact that `everybody knows exactly her own
secret'. (ii) In general, each agent always considers it possible that
call sequences (of any length) take place that do not involve her.
These assumptions are weaker than the ones analyzed in \cite{ADGH14}.

\smallskip

We state without proof the following simple fact.
\begin{fact}
\NI
\begin{enumerate}[(i)]
\item Each $\sim_a$ is an equivalence relation;
\item For all $\CSequence, \CSequenced \in \CSequences$ if $\CSequence \sim_a \CSequenced$, then $\CSequence(\init)_a = \CSequenced(\init)_a$, but not vice versa.
\end{enumerate}
\end{fact}
This prompts us to note also that according to Definition \ref{def:model} sequences which make $a$ learn the same set of secrets may well be distinguishable for $a$, such as, for instance, $ab,bc,ab$ and $ab,bc,ac$. In the first one $a$ comes to know that $b$ knows $a$ is familiar with all secrets, while in the second one, she comes to know that $c$ knows $a$ is familiar with all secrets. Relation $\sim_a$ is so defined as to capture this sort of `higher-order' knowledge.

\subsubsection{Truth conditions for epistemic guards}

Everything is now in place to define the truth of the considered formulas.
\begin{definition}
Let $(\Model, \CSequence)$ be a pointed gossip model with $\Model =
(\CSequences^{< \omega}, (\sim_a)_{a \in \Agents})$ and $\CSequence \in
\CSequences^{< \omega}$. We define the satisfaction relation $\models$
inductively as follows (clauses for Boolean connectives are omitted):
\begin{eqnarray*}
(\Model, \CSequence) \models F_a p & \mbox{iff} & p \in \CSequence(\init)_a, \\
(\Model, \CSequence) \models K_a \phi &  \mbox{iff}  & \forall \CSequenced \mbox{     s.t.     } \CSequence \sim_a \CSequenced, ~(\Model, \CSequenced) \models \phi. 
\end{eqnarray*}
\end{definition}
So formula $F_a p$ is true (in a pointed gossip model)
whenever secret $p$ belongs to the set of secrets
agent $a$ is familiar with in the situation generated by the
designated call sequence $\CSequence$ applied to the initial situation
$\init$. The knowledge operator is interpreted as customary in epistemic logic using the
equivalence relations $\sim_a$.



\medskip

\subsubsection{Computations}

Assume a gossip protocol $P$ that is a parallel composition of the component programs
$* [[]^{m_a}_{j=1}\ \psi^a_j \to \Call^a_j]$, one for each agent $a \in \Agents$.

Given the gossip model $\Model = (\CSequences^{< \omega},
\set{\sim_a}_{a \in \Agents})$ we define the \bfe{computation tree}
$\CSequences^P \subseteq \CSequences^{< \omega}$ of $P$ as the
smallest set of sequences satisfying the following inductive
conditions:
\begin{description}
\item{[Base]} $\epsilon \in \CSequences^P$; 
\item{[Step]} If $\CSequence \in  \CSequences^P$ and $(\Model, \CSequence) \models \psi^a_j$ then $\CSequence.\Call^a_j \in  \CSequences^P$. In this case we say that a \bfe{transition} has taken place between $\CSequence$ and $\CSequence.\Call^a_j$, in symbols, $\CSequence \to \CSequence.\Call^a_j$.
\end{description}
So $\CSequences^P$ is a (possibly infinite) set of finite call
sequences that is iteratively obtained by performing a
`legal' call (according to protocol $P$) from a `legal' (according to
protocol $P$) call sequence.

A \bfe{path} in the computation tree of $P$ is a (possibly infinite)
sequence of elements of $\CSequences^P$, denoted by $\xi =
(\CSequence_0, \CSequence_1, \ldots, \CSequence_n, \ldots)$, where
$\CSequence_0 = \epsilon$ and each $\CSequence_{i+1} =
\CSequence_i.\Call$ for some call $\Call$ and $i \geq 0$. A
\bfe{computation} of $P$ is a maximal rooted path in the computation tree of
$P$.\footnote{Note that while the sequences that are elements of the
  computation tree of a protocol are always finite (although possibly
  infinite in number), computations can be infinite sequences (of
  finite call sequences).}


\smallskip

The above definition implies that a call sequence
$\CSequence$ is a leaf of the computation tree if and only if
\[
(\Model, \CSequence) \models \bigwedge_{a \in \mathsf{A}}
\bigwedge^{m_a}_{j=1} \ \neg \psi^a_j.
\]
We call the formula 
\[
\bigwedge_{a \in \mathsf{A}} \bigwedge^{m_a}_{j=1} \ \neg \psi^a_j
\]
the \bfe{exit condition} of the gossip protocol $P$.

\smallskip

Obviously computation trees can be infinite, though they are always
finitely branching.  Further, note that this semantics for gossip protocols abstracts away from
some implementation details of the calls.  More specifically, we
assume that the caller always succeeds in his call and does not
require to synchronize with the called agent.  In reality, the called
agent might be busy, being engaged in another call.  To take care of
this one could modify each call by replacing it by a `call protocol'
that implements the actual call using some lower level primitives. 
We do not elaborate further on this topic.

\medskip


Let us fix some more terminology. For $\CSequence \in \CSequences^P$, an agent $a$ is \bfe{enabled} in $\CSequence$ if $(\Model, \CSequence)
\models \bigvee^{m_a}_{j=1} \ \psi^a_j$ and is \bfe{disabled}
otherwise. So an agent is enabled if it can perform a call.  An agent
$a$ is \bfe{selected} in $\CSequence$ if it is the caller in the call
that for some $\CSequence'$ determines the transition $\CSequence \to \CSequence'$
in $\xi$.  Finally, a computation $\xi$ is called a \bfe{fair computation} if it is
finite or each agent that is enabled in infinitely many
sequences in $\xi$ is selected in infinitely many sequences
in $\xi$.

We note in passing that various alternative definitions of fairness
are possible; we just focus on one of them. An interested reader may
consult \cite{AFK88}, where several fairness definitions (for instance
one focusing on actions and not on agents) for distributed programs
were considered and compared.

\smallskip

We conclude this section by observing the following.  Our definition
of computation tree for protocol $P$ presupposes that guards
$\psi^a_j$ are interpreted over the gossip model $\Model =
(\CSequences^{< \omega}, \set{\sim_a}_{a \in \Agents})$. This means
that when evaluating guards, agents consider as possible call
sequences that cannot be generated by $P$. In other words, agents do
not know the protocol. To model common knowledge of the considered
protocol in the gossip model one should take as the domain of the
gossip model $\Model$ the underlying computation tree.  However, the
computation tree is defined by means of the underlying gossip model.
To handle such a circularity an appropriate fixpoint definition is
needed.  We leave this topic for future work.


\subsection{Correctness}

We are interested in proving the correctness of gossip protocols. 
Assume a gossip protocol $P$ that is a parallel composition of the component programs
$* [[]^{m_a}_{j=1}\ \psi^a_j \to \Call^a_j]$. 

We say that $P$ is \bfe{partially correct}, in short \bfe{correct}, if 
in all situations sequences $\CSequence$
that are leaves of the computation tree of $P$, for each agent $a$ 
\[
(\Model, \CSequence) \models \bigwedge_{b \in \mathsf{A}} F_a B,
\]
i.e., if for all situations sequences $\CSequence$
that are leaves of the computation tree of $P$, each agent is
an expert in the gossip situation $\CSequence(\init)$.


We say furthermore that $P$ \bfe{terminates} if all its computations are finite
and that $P$ \bfe{fairly terminates} if all its fair computations are finite.

\smallskip

In the next section we provide examples showing that partial
correctness and termination of the considered protocols can depend on
the assumed mode of communication and on the number of agents.
%
In what follows we study various gossip protocols and their
correctness. We begin with the following obvious observation.

\begin{fact}\label{not:1}
  For each protocol $P$ the following implications ($\Rightarrow$) hold,
  where $T_P(x)$ stands for its termination and $FT_P(x)$ for its fair
  termination in a communication mode $x$:
\[
T_P(x) \Rightarrow FT_P(x).
\]
\end{fact}

Protocol R3 given in Section \ref{sec:ring} shows that none of these
implications can be reversed.  Moreover, it is not the case either that for each protocol $P$:
\begin{align*}
T_P(\push) \Rightarrow T_P(\mbox{push-pull}), \\ 
T_P(\pull) \Rightarrow T_P(\mbox{push-pull}). 
\end{align*}

\begin{example}
\rm
Let  $\Agents = \set{a, b, c}$ and define the following expression:
\begin{eqnarray*}
\mathcal{A} \subset \mathcal{C} & := & \bigwedge_{I \in \{A,B,C\}} (F_a I \to F_c I) \land 
\bigvee_{I \in \{A,B,C\}} (F_c I \land \neg F_a I)
\end{eqnarray*}
Expression $\mathcal{B} \subset \mathcal{C}$ is defined analogously. Note that we denote by $I$ the secret of agent $i$.
Intuitively, $\mathcal{A} \subset \mathcal{C}$ means that agent $c$
is familiar with all the secrets that agent $a$ is familiar with, but not vice versa. 
So $c$ is familiar with a strict superset of the secrets $a$ is familiar with.
Further, let $Exp_j$ stand for $\bigwedge_{I \in \{A,B,C\}} F_j I$.

Consider now the following component programs:

\begin{itemize}

\item for agent $a$: 
$
*[\neg K_a (\mathcal{A} \subset \mathcal{C}) \land \neg K_a Exp_a  \to a \push c],
$

\item for agent $b$:
$
*[\neg K_b (\mathcal{B} \subset \mathcal{C}) \land \neg K_b Exp_b  \to b \push c],
$

\item for agent $c$: 
$
*[\neg K_c Exp_a \land K_c Exp_c  \to c \push a \: [] \: \neg K_c Exp_b \land K_c Exp_c  \to c \push b].
$
\end{itemize}

This protocol is correct. Indeed, initially no agent is an expert,
hence both guards of $c$ are false.  On the other hand, we have
$(\Model, \epsilon) \models \neg (\mathcal{A} \subset \mathcal{C})$ and
$(\Model, \epsilon) \models \neg (\mathcal{B} \subset \mathcal{C})$, so both
$(\Model, \epsilon) \models \neg K_a (\mathcal{A} \subset \mathcal{C})$ and
$(\Model, \epsilon) \models \neg K_b (\mathcal{B} \subset \mathcal{C})$.
Consequently, initially both $a$ and $b$ are enabled.  If the first
call is granted to $a$, this agent will call $c$ yielding the gossip
situation $A.B.AC$. Now the guard of $a$ is false (since $a$ is still
familiar only with his own secret $A$, while $c$ is familiar with at
least $A$ and $C$ and $a$ knows this).  The guard of $c$ is still
false. So now only $b$ is enabled. After his call of $c$ this yields
the gossip situation $A.B.ABC$. At this stage, only agent $c$ is
enabled and after he calls both $a$ and $b$ all guards become false.
Moreover, this protocol terminates. Indeed, the only computations are
the ones in which first the calls $a \push c$ and $b \push c$ take
place, in any order, followed by the calls $c \push a$ and
$c \push b$, also performed in any order.  

However, if we use the push-pull direction type instead of push, then
the situation changes.  Indeed, after an arbitrary number of calls
$ac$ the formula $\neg (\mathcal{A} \subset \mathcal{C})$ is still
true and hence $\neg K_a (\mathcal{A} \subset \mathcal{C})$ is true,
as well.  Consequently, this call can be indefinitely repeated, so the
protocol does not terminate. 
\HB
\end{example}

\section{Two symmetric protocols} \label{section:complete}
\label{sec:complete}

In this section we consider protocols for the case when the agents
form a complete graph.  We study two protocols. We present
them first for the communication mode push-pull.  (Partial) correctness of the considered protocols does not depend on
the assumed mode of communication.

\paragraph{Learn new secrets protocol (LNS)}
\NI
Consider the following program for agent $i$:
\[
*[[]_{j \in \mathsf{A}} \neg F_i J \to (i, j)].
\]
Informally, agent $i$ calls agent $j$ if $i$ is not familiar with
$j$'s secret. Note that the guards of this protocol do not use the
epistemic operator $K_i$, but they are equivalent to the ones that do, as
$\neg F_iJ$ is equivalent to $K_i\neg F_iJ$.

This protocol was introduced in \cite{ADGH14} and studied with respect
to the push-pull mode, assuming asynchronous communication. As noted
there this protocol is clearly correct.  Also, it always terminates
since after each call $(i, j)$ the size of $\set{(i,j)\in \Agents
  \times \Agents \mid \neg F_i J}$ decreases.  The same argument shows
termination if the communication mode is pull.

However, if the communication mode is push, the protocol may fail to
terminate, even fairly.  To see it fix an agent $a$ and consider a
sequence of calls in which each agent calls $a$.  At the end of this
sequence $a$ becomes an expert but nobody is familiar with his secret.
So any extension of this sequence is an infinite computation.

\smallskip

Let us consider now the possible call sequences generated by the
computations of this protocol.  Assume that there are $n \geq 4$
agents. By the result mentioned in the introduction in each
terminating computation at least $2n - 4$ calls are made.

The LNS protocol can generate such shortest sequences (among others).
Indeed, let $\mathsf{A} = \{a,b,c,d,$ $i_1, $ $\LL, i_{n-4}\}$ be the set of
agents.  Then the following sequence of $2n - 4$ calls
\begin{equation}\label{eq:shortest sequence}
\begin{array}{c}
(a,i_1), (a,i_2), \LL, (a,i_{n-4}),\\ (a,b), (c,d), (a,c), (b,d),\\
 (i_1,b),
(i_2,b), \LL, (i_{n-4},b)
\end{array}
\end{equation}
corresponds to a terminating computation.

The guards used in this protocol entail that after a call $(i,j)$
neither the call $(j,i)$ nor another call $(i,j)$ can take place, that
is between each pair of agents at most one call can take place.
Consequently, the longest possible sequence contains at most $\frac{n(n-1)}{2}$ calls.  Such a worst case can be generated by means of
the following sequence of calls:
\[
[2], \ [3], \ [4], \LL, [n],
\]
where for a natural number $k$, $[k]$ stands for the sequence
$(1,k),$ $(2,k),$ $\LL,$ $(k-1,k)$.\footnote{Other longest sequences are obviously possible, for instance: $12,13,...,1n, 23,24,...,2n, 34,35,..,3n,  ... , (n-1)n$.}

\paragraph{Hear my secret protocol (HMS)}

\NI
Next, we consider a protocol with the following program for agent $i$:
\[
*[[]_{j \in \mathsf{A}} \neg K_i F_j I \to (i, j)].
\]
Informally, agent $i$ calls agent $j$ if he (agent $i$) does not know whether $j$
is familiar with his secret. 
To prove correctness of this protocol it suffices to note that its exit condition 
\[
\bigwedge_{i,j \in \mathsf{A}} K_i F_j I
\]
implies $\bigwedge_{i,j \in \mathsf{A}} F_j I$. To prove termination it suffices to note that after each call $(i, j)$ the size of the set
$\{(i,j) \mid \neg K_i F_j I\}$ decreases.

\smallskip

If the communication mode is push, then the termination argument
remains valid, since after the call $i \push j$ agent $j$ still learns
all the secrets agent $i$ is familiar with.

However, if the communication mode is pull, then the protocol may fail
to terminate, even fairly. To see it fix an agent $j$ and consider the
calls $i \pull j$, where $i$ ranges over $\mathsf{A} \setminus \{j\}$,
arbitrarily ordered.  Denote this sequence by $\CSequence$.  Consider
now an infinite sequence of calls resulting from repeating $\CSequence$
indefinitely. It is straightforward to check that such a sequence
corresponds to a possible computation.  Indeed, in this sequence agent
$j$ never calls and hence never learns any new secret.  So for each $i
\neq j$ the formula $\neg K_i F_j I$ remains true and hence each
agent $i \neq j$ remains enabled.  Moreover, after the calls from
$\CSequence$ took place agent $j$ is not anymore enabled.  Hence the
resulting infinite computation is fair.


\section{Protocols over directed rings} \label{section:rings}
\label{sec:ring}

In this section we consider the case when the agents are arranged in a
directed ring, where $n \geq 3$. For convenience we take the set of
agents to be $\set{1, 2, \LL, n}$.  For $i \in \{1, \LL, n\}$, let $i
\oplus 1$ and $i \ominus 1$ denote respectively the successor and
predecessor of agent $i$.  That is, for $i \in \C{1,\LL,n-1}$, $i
\oplus 1=i+1$, $n \oplus 1=1$, for $i \in \C{2,\LL,n}$, $i \ominus
1=i-1$, and $1 \ominus 1=n$. For $k > 1$ we define $i \oplus k$ and $i \ominus k$
by induction in the expected way.
Again, when reasoning about the protocols we
denote the secret of agent $i \in \{1, \LL, n\}$ by $I$.
We consider four different protocols and study them with respect to
their correctness and (fair) termination.

In this set up, a call sequence over a directed ring is a (possibly
infinite) sequence of calls, all being of the same communication mode,
and all involving an agent $i$ and $i \oplus 1$.  As before, we use
$\CSequence$ to denote such a call sequence and $\CSequences_{DR}$ to
denote the set of all call sequences over a directed ring.  In this
section, unless stated otherwise, by a call sequence we mean a
sequence over a directed ring.  The set of all such finite call
sequences is denoted $\CSequences_{DR}^{< \omega}$.  A gossip model
for a directed ring is a tuple $\Model_{DR} = (\CSequences_{DR}^{<
  \omega}, \set{\sim_a}_{a \in \Agents})$, where each $\sim_a
\subseteq \CSequences_{DR}^{< \omega} \times \CSequences_{DR}^{<
  \omega}$ is as in Definition~\ref{def:model}. The truth definition
is as before, and the notion of a \bfe{computation tree for directed
  rings} $\CSequences_{DR}^P \subseteq \CSequences_{DR}^{< \omega}$ of
a ring protocol $P$ is analogous to the notion defined before. Note that by
restricting the domain in $\Model_{DR}$ to $\CSequences_{DR}^{<
  \omega}$, the ring network---and hence who is the successor of whom---becomes common knowledge.

When presenting the protocols we use the fact that $F_iJ$ is
equivalent to $K_i F_iJ$.

\paragraph{Ring protocol R1}
\NI
Consider first a gossip protocol with the following program for $i$:
\[
*[\bigvee_{j=1}^{n} (F_i J \land K_i \neg F_{i \oplus 1} J) \to i \Diamond i\oplus 1],
\]
where $\Diamond$ denotes the mode of communication, so $\push$, $\pull$ or push-pull.

Informally, agent $i$ calls his successor, agent $i \oplus 1$, if
$i$ is familiar with some secret and he knows that his
successor is not familiar with it.

\begin{proposition} \label{proposition:R1} 
Let $\Diamond = \push$. Protocol R1 terminates and is correct.
\end{proposition}

Termination and correctness do not both hold for the other communication modes.
Consider first the pull communication mode, i.e., $\Diamond = \pull$.
Then the protocol does not always terminate. 
Indeed, each call $i \pull i \oplus 1$ can be repeated.
Next, consider the push-pull communication mode.
We show that then the protocol is not correct. Indeed, take
\[
\CSequence = (1,2), \ (2, 3), \LL, (n-1, n).
\]
We claim that after the sequence of calls $\CSequence$ the exit condition
of the protocol is true.  To this end we consider each agent in turn.

After $\CSequence$ each agent $i$, where $i \neq n$ is familiar the secrets of the agents $1, 2, \LL, i+1$.
Moreover, because of the call $(i,i+1)$ agent $i$ knows that agent $i+1$ is familiar with these secrets.
So the exit condition of agent $i$ is true.

To deal with agent $n$ note that $\CSequence \sim_n \CSequence. (n-2,
n-1). (n-3, n-2). \LL (2, 3). (1,2)$. After the latter call sequence
agent 1 becomes an expert. So after $\CSequence$ agent $n$
cannot know that agent 1 is not familiar with some secret.
Consequently, after $\CSequence$ the exit condition of agent $n$ is
true, as well. However, after $\CSequence$ agent $1$ is not an expert,
so the protocol is indeed not correct.

\bigskip

In what follows we initially present the protocols assuming the push-pull
mode of communication.

\paragraph{Ring protocol R2}
\NI
Consider now a gossip protocol with the following program for agent $i$:
\[
*[\neg K_i F_{i \oplus 1} I \ominus 1 \to (i, i \oplus 1)],
\]
where (recall) $I \ominus 1$ denotes the secret of agent $i \ominus 1$.
Informally, agent $i$ calls his successor, which is agent $i \oplus 1$, if $i$
does not know that his successor is familiar with the secret of $i$'s
predecessor, i.e., agent $i \ominus 1$.

\begin{proposition} \label{proposition:R2}
If $|\Agents| \in \set{3,4}$ then protocol R2 is correct.
\end{proposition}

However, this protocol is not correct for five or more agents. To see it
consider the sequence of calls
\[
(1,2), \ (2,3), \ \LL, (n-1, n), \ (n,1), \ (1,2)
\]
where $n \geq 5$.  After it the exit condition of the protocol is
true. However, agent 3 is not familiar with the secret of agent 5.


Note that the same argument shows that the protocol in which we use
$\neg K_i F _{i \oplus 1} I \vee \neg K_i F_{i \oplus 1} I \ominus
1$ instead of $\neg K_i F_{i \oplus 1} I \ominus 1$ is incorrect, as well.

Moreover, this protocol does not always terminate.
Indeed, one possible computation consists of an agent $i$
repeatedly calling his successor $i \oplus 1$.

\paragraph{Ring protocol R3}
\NI
Next, consider the following modification of protocol R2
in which we use the following program for agent $i$:
\[
*[(\neg \bigwedge_{j = 1}^{n} F_i J) \vee \neg K_i F_{i \oplus
  1} I \ominus 1 \to (i, i \oplus 1)].
\]
Informally, agent $i$ calls his successor, agent $i \oplus 1$, if
$i$ is not familiar with all the secrets or $i$ does not know that his
successor is familiar with the secret of his predecessor, agent $i
\ominus 1$. 

This gossip protocol is obviously correct thanks to the fact that
$\bigwedge_{i = 1}^{n} \bigwedge_{j = 1}^{n} F_i J$ is part of the
exit condition.  However, it does not always
terminate for the same reason as the previous one.

On the other hand, the following holds.
\begin{proposition} \label{proposition:R3}
Protocol R3 fairly terminates.
\end{proposition}

The same conclusions concerning non termination and fair termination
can be drawn for the push and the pull modes of communication. Indeed, for
push it suffices to consider the sequence of calls $i \push i \oplus
1, \ i \oplus 1 \push i \oplus 2, \LL, i \ominus 1 \push i$ after
which agent $i \ominus 1$ becomes disabled, and for pull the sequence
of calls $i \pull i \oplus 1, \ i \ominus 1 \pull i, \LL, i \oplus 2
\pull i \oplus 3$ after which agent $i \oplus 2$ becomes disabled.

\paragraph{Ring protocol R4}
\NI
Finally, we consider a protocol that is both correct and terminates for
the push-pull mode. Consider the following program for $i$:
\[
*[\bigvee_{j=1}^{n} (F_i J \wedge \neg K_i F_{i \oplus 1} J) \to (i, i \oplus 1)].
\]
Informally, agent $i$ calls his successor, agent $i \oplus 1$, if
$i$ is familiar with some secret and he does not know whether his
successor is familiar with it. Note the similarity with protocol R1.

\begin{proposition} \label{proposition:R4}
Protocol R4 terminates and is correct.
\end{proposition}

If the communication mode is push, then the termination argument
remains valid, since after the call $i \push i \oplus 1$ agent $i
\oplus 1$ still learns all the secrets that agent $i$ is familiar with
and hence the above set $\{(i,j) \mid \neg K_i F_{i \oplus 1} J \}$
decreases.

If the communication mode is pull, then the protocol may fail to
terminate, because after the first call $i \pull i \oplus 1$ agent $i
\oplus 1$ does not learn the secret of agent $i$ and consequently the
call can be repeated. However, the situation changes when fairness is
assumed.

\begin{proposition} \label{proposition:R4f}
For the pull communication mode protocol R4 fairly terminates.
\end{proposition}

\smallskip

Table \ref{figure:termination} summarizes the termination properties of the protocols
considered in the paper.

\begin{table}
\begin{center}
\begin{tabular}{ c  c  c  c  c  c  c }
\hline
Protocol & T   & FT  & T for $\push$ & FT for $\push$ & T for $\pull$ &FT for $\pull$ \\
\hline
LNS       & yes  &yes  &no      &no  &yes          &yes        \\
HMS        & yes  &yes  &yes      &yes  &no          &no        \\
R3        & no  &yes  &no      &yes  &no          &yes        \\
R4        & yes  &yes  &yes      &yes  &no          &yes        \\
\hline
\end{tabular}
\end{center}
\caption{Summary of termination results.}
\label{figure:termination}
\end{table}
\section{Conclusions}

The aim of this paper was to introduce distributed gossip
protocols, to set up a formal framework to reason about them, and to
illustrate it by means of an analysis of selected protocols. 

\smallskip

Our results open up several avenues for further research.  First, our
correctness arguments were given in plain English with occasional
references to epistemic tautologies, such as $K_i \phi \to \phi$, but
it should be possible to formalize them in a customized epistemic
logic. Such a logic should have a protocol independent component that
would consist of the customary S5 axioms and a protocol dependent
component that would provide axioms that depend on the mode of
communication and the protocol in question.  An example of such an
axiom is the formula $K_i F_{i \oplus 1} I \ominus 1 \to F_i I \oplus
1$ that we used when reasoning about protocol R2. To prove the
validity of the latter axioms one would need to develop a proof system
that allows us to compute the effect of the calls, much like the
computation of the strongest postconditions in Hoare logics.
Once such a logic is provided the next step will be to study formally
its properties, including decidability. Then we could clarify
whether the provided correctness proofs could be carried out automatically.

Second, generalizing further the ideas we introduced by considering
directed rings, gossip protocols could be studied in interface with
network theory (see \cite{jackson08social} for a textbook
presentation).  Calls can be assumed to be constrained by a network,
much like in the literature on `centralized' gossip (cf. \cite{HHL88})
or even have probabilistic results (i.e., secrets are passed with
given probabilities). More complex properties of gossip protocols
could then be studied involving higher-order knowledge or forms of group knowledge among
neighbors (e.g., ``it is common knowledge among $a$ and her neighbors
that they are all experts''), or their stochastic behavior (e.g., ``at
some point in the future all agents are experts with probability
$p$'').


Third, it will be interesting to analyze the protocols for the types of calls
considered in  \cite{ADGH14}. They presuppose some form of knowledge
that a call took place (for instance that given a call between $a$ and $b$
each agent $c \neq a,b$ noted the call but did not learn its content).
Another option is to consider multicasting (calling several agents at the
same time).

Finally, many assumptions of the current setup could be lifted.
Different initial and final situations could be considered, for
instance common knowledge of protocols could be assumed, or common
knowledge of the familiarity of all agents with all the secrets upon
termination could be required. Finally, to make the protocols more
efficient passing of tokens could be allowed instead of just the
transmission of secrets by means of calls.


\subsection*{Acknowledgments}

We would like to thank Hans van Ditmarsch and the referees for helpful
comments and Rahim Ramezanian for useful comments about Example 2.8.
This work resulted from a research visit by Krzysztof Apt to Davide
Grossi and Wiebe van der Hoek, sponsored by the 2014 Visiting
Fellowship Scheme of the Department of Computer Science of the
University of Liverpool.  The first author is also a Visiting Professor at
the University of Warsaw.  He was partially supported by the NCN grant
nr 2014/13/B/ST6/01807.

\nocite{*}
\bibliographystyle{eptcs}
\bibliography{../gossip_TARK_cameraready_amended/distributed_gossip}

\begin{thebibliography}{10}
\providecommand{\bibitemdeclare}[2]{}
\providecommand{\surnamestart}{}
\providecommand{\surnameend}{}
\providecommand{\urlprefix}{Available at }
\providecommand{\url}[1]{\texttt{#1}}
\providecommand{\href}[2]{\texttt{#2}}
\providecommand{\urlalt}[2]{\href{#1}{#2}}
\providecommand{\doi}[1]{doi:\urlalt{http://dx.doi.org/#1}{#1}}
\providecommand{\bibinfo}[2]{#2}

\bibitemdeclare{book}{ABO09}
\bibitem{ABO09}
\bibinfo{author}{K.~R. \surnamestart Apt\surnameend}, \bibinfo{author}{F.~R.
  \surnamestart {de Boer}\surnameend} \& \bibinfo{author}{E.~R. \surnamestart
  Olderog\surnameend} (\bibinfo{year}{2009}):
  \emph{\bibinfo{title}{Verification of Sequential and Concurrent Programs}}.
\newblock \bibinfo{publisher}{Springer}, \doi{10.1007/978-1-84882-745-5}.

\bibitemdeclare{article}{AFK88}
\bibitem{AFK88}
\bibinfo{author}{K.~R. \surnamestart Apt\surnameend},
  \bibinfo{author}{N.~\surnamestart Francez\surnameend} \&
  \bibinfo{author}{S.~\surnamestart Katz\surnameend} (\bibinfo{year}{1988}):
  \emph{\bibinfo{title}{Appraising fairness in distributed languages}}.
\newblock {\sl \bibinfo{journal}{Distributed Computing}}
  \bibinfo{volume}{2}(\bibinfo{number}{4}), pp. \bibinfo{pages}{226--241},
  \doi{10.1007/BF01872848}.

\bibitemdeclare{inproceedings}{ADGH14}
\bibitem{ADGH14}
\bibinfo{author}{M.~\surnamestart Attamah\surnameend},
  \bibinfo{author}{H.~\surnamestart {van Ditmarsch}\surnameend},
  \bibinfo{author}{D.~\surnamestart Grossi\surnameend} \&
  \bibinfo{author}{W.~\surnamestart {Van der Hoek}\surnameend}
  (\bibinfo{year}{2014}): \emph{\bibinfo{title}{Knowledge and gossip}}.
\newblock In: {\sl \bibinfo{booktitle}{Proceedings of ECAI'14}},
  \bibinfo{publisher}{IOS Press}, pp. \bibinfo{pages}{21--26}.

\bibitemdeclare{article}{baker72gossips}
\bibitem{baker72gossips}
\bibinfo{author}{B.~\surnamestart Baker\surnameend} \&
  \bibinfo{author}{R.~\surnamestart Shostak\surnameend} (\bibinfo{year}{1972}):
  \emph{\bibinfo{title}{Gossips and Telephones}}.
\newblock {\sl \bibinfo{journal}{Discrete Mathematics}} \bibinfo{volume}{2},
  pp. \bibinfo{pages}{197--193}, \doi{10.1016/0012-365X(72)90001-5}.

\bibitemdeclare{article}{bumby81problem}
\bibitem{bumby81problem}
\bibinfo{author}{R.~\surnamestart Bumby\surnameend} (\bibinfo{year}{1981}):
  \emph{\bibinfo{title}{A Problem with Telephones}}.
\newblock {\sl \bibinfo{journal}{SIAM Journal of Algorithms and Discrete
  Methods}} \bibinfo{volume}{2}, pp. \bibinfo{pages}{13--18},
  \doi{10.1137/0602002}.

\bibitemdeclare{article}{Dij75}
\bibitem{Dij75}
\bibinfo{author}{E.~W. \surnamestart Dijkstra\surnameend}
  (\bibinfo{year}{1975}): \emph{\bibinfo{title}{Guarded commands,
  nondeterminacy and formal derivation of programs}}.
\newblock {\sl \bibinfo{journal}{Communications of the {ACM}}}
  \bibinfo{volume}{18}, pp. \bibinfo{pages}{453--457},
  \doi{10.1145/360933.360975}.

\bibitemdeclare{article}{fagin97knowledge}
\bibitem{fagin97knowledge}
\bibinfo{author}{R.~\surnamestart Fagin\surnameend},
  \bibinfo{author}{J.~\surnamestart Halpern\surnameend},
  \bibinfo{author}{Y.~\surnamestart Moses\surnameend} \&
  \bibinfo{author}{M.~\surnamestart Vardi\surnameend} (\bibinfo{year}{1997}):
  \emph{\bibinfo{title}{Knowledge-Based Programs}}.
\newblock {\sl \bibinfo{journal}{Distributed Computing}} \bibinfo{volume}{10},
  pp. \bibinfo{pages}{199--225}, \doi{10.1007/s004460050038}.

\bibitemdeclare{book}{fagin95reasoning}
\bibitem{fagin95reasoning}
\bibinfo{author}{Ronald \surnamestart Fagin\surnameend},
  \bibinfo{author}{Joseph~Y. \surnamestart Halpern\surnameend},
  \bibinfo{author}{Yoram \surnamestart Moses\surnameend} \&
  \bibinfo{author}{Moshe~Y. \surnamestart Vardi\surnameend}
  (\bibinfo{year}{1995}): \emph{\bibinfo{title}{Reasoning about knowledge}}.
\newblock \bibinfo{publisher}{The MIT Press}, \bibinfo{address}{Cambridge}.

\bibitemdeclare{article}{hajnal72cure}
\bibitem{hajnal72cure}
\bibinfo{author}{A.~\surnamestart Hajnal\surnameend}, \bibinfo{author}{E.~C.
  \surnamestart Milner\surnameend} \& \bibinfo{author}{E.~\surnamestart
  Szemeredi\surnameend} (\bibinfo{year}{1972}): \emph{\bibinfo{title}{A Cure
  for the Telephone Disease}}.
\newblock {\sl \bibinfo{journal}{Canadian Mathematical Bulletin}}
  \bibinfo{volume}{15}, pp. \bibinfo{pages}{447--450},
  \doi{10.4153/CMB-1972-081-0}.

\bibitemdeclare{article}{HHL88}
\bibitem{HHL88}
\bibinfo{author}{S.~M. \surnamestart Hedetniemi\surnameend},
  \bibinfo{author}{S.~T. \surnamestart Hedetniemi\surnameend} \&
  \bibinfo{author}{A.~L. \surnamestart Liestman\surnameend}
  (\bibinfo{year}{1988}): \emph{\bibinfo{title}{A survey of gossiping and
  broadcasting in communication networks}}.
\newblock {\sl \bibinfo{journal}{Networks}}
  \bibinfo{volume}{18}(\bibinfo{number}{4}), pp. \bibinfo{pages}{319--349},
  \doi{10.1002/net.3230180406}.

\bibitemdeclare{article}{Hoa78}
\bibitem{Hoa78}
\bibinfo{author}{C.~A.~R. \surnamestart Hoare\surnameend}
  (\bibinfo{year}{1978}): \emph{\bibinfo{title}{Communicating sequential
  processes}}.
\newblock {\sl \bibinfo{journal}{Communications of the {ACM}}}
  \bibinfo{volume}{21}, pp. \bibinfo{pages}{666--677},
  \doi{10.1145/359576.359585}.

\bibitemdeclare{book}{INM84}
\bibitem{INM84}
\bibinfo{author}{\surnamestart {INMOS Limited}\surnameend}
  (\bibinfo{year}{1984}): \emph{\bibinfo{title}{Occam Programming Manual}}.
\newblock \bibinfo{publisher}{Prentice-Hall International}.

\bibitemdeclare{book}{jackson08social}
\bibitem{jackson08social}
\bibinfo{author}{M.~O. \surnamestart Jackson\surnameend}
  (\bibinfo{year}{2008}): \emph{\bibinfo{title}{Social and Economic Networks.}}
\newblock \bibinfo{publisher}{Princeton University Press}.

\bibitemdeclare{inproceedings}{kurki-suonio86towards}
\bibitem{kurki-suonio86towards}
\bibinfo{author}{R.~\surnamestart Kurki-Suonio\surnameend}
  (\bibinfo{year}{1986}): \emph{\bibinfo{title}{Towards programming with
  Knowledge Expressions}}.
\newblock In: {\sl \bibinfo{booktitle}{Proceedings of POPL'86}}, pp.
  \bibinfo{pages}{140--149}, \doi{10.1145/512644.512657}.

\bibitemdeclare{book}{meyer95epistemic}
\bibitem{meyer95epistemic}
\bibinfo{author}{{J.-J. Ch.} \surnamestart Meyer\surnameend} \&
  \bibinfo{author}{W.~\surnamestart van~der Hoek\surnameend}
  (\bibinfo{year}{1995}): \emph{\bibinfo{title}{Epistemic Logic for AI and
  Computer Science}}.
\newblock {\sl \bibinfo{series}{Cambridge Tracts in Theoretical Computer
  Science}}~\bibinfo{volume}{41}, \bibinfo{publisher}{Cambridge University
  Press}, \doi{10.1017/CBO9780511569852}.

\bibitemdeclare{inproceedings}{parikhetal:1985}
\bibitem{parikhetal:1985}
\bibinfo{author}{R.~\surnamestart Parikh\surnameend} \&
  \bibinfo{author}{R.~\surnamestart Ramanujam\surnameend}
  (\bibinfo{year}{1985}): \emph{\bibinfo{title}{Distributed Processing and the
  Logic of Knowledge}}.
\newblock In: {\sl \bibinfo{booktitle}{Logic of Programs}},
  \bibinfo{series}{LNCS 193}, \bibinfo{publisher}{Springer}, pp.
  \bibinfo{pages}{256--268}, \doi{10.1007/3-540-15648-8}.
\newblock \bibinfo{note}{Similar to {\em JoLLI} 12: 453--467, 2003}.

\bibitemdeclare{article}{seress86quick}
\bibitem{seress86quick}
\bibinfo{author}{\'{A} \surnamestart Seress\surnameend} (\bibinfo{year}{1986}):
  \emph{\bibinfo{title}{Quick Gossiping without Duplicate Transmissions}}.
\newblock {\sl \bibinfo{journal}{Graphs and Combinatorics}}
  \bibinfo{volume}{2}, pp. \bibinfo{pages}{363--383}, \doi{10.1007/BF01788111}.

\bibitemdeclare{article}{tijdeman:1971}
\bibitem{tijdeman:1971}
\bibinfo{author}{R.~\surnamestart Tijdeman\surnameend} (\bibinfo{year}{1971}):
  \emph{\bibinfo{title}{On a telephone problem}}.
\newblock {\sl \bibinfo{journal}{Nieuw Archief voor Wiskunde}}
  \bibinfo{volume}{3(XIX)}, pp. \bibinfo{pages}{188--192}.

\end{thebibliography}

\appendix

\vspace{0.5cm}

\begin{proof}[of Proposition \ref{proposition:R1}]
\mbox{}

\NI
\fbox{Termination}
Given a call sequence $\CSequence$  define the set
\[
\mathit{Inf}(\CSequence) := \{ (i,j) \mid i,j \in \{1, \LL, n\} \mbox{   and   }
(\Model_{DR}, \CSequence) \models F_i J \}.
\]
After each enabled call $i \push i \oplus 1$ in $\CSequence$,  the set $\mathit{Inf}(\CSequence)$ increases, which
ensures termination since each set $\mathit{Inf}(\cdot)$ has at most $n^2$ elements.

\NI
\fbox{Correctness}
Consider a leaf of the computation tree. Then the exit condition
\[
\bigwedge_{i=1}^{n} \bigwedge_{j=1}^{n} (\neg F_i J \vee \neg K_i \neg F_{i \oplus 1} J)
\]
is true. We proceed by induction to show that then each $F_iJ$ is true, where $i, j \in \{1, \LL,
n\}$, and where the pairs $(i,j)$ are ordered as follows:
\begin{align*}
& (1,1), (2,1), \LL, (n,1), \\ 
& (2,2), (3,2), \LL, (1,2), \\
& \LL, \\
& (n,n), (1,n), \LL, (n-1,n).
\end{align*}
So the $i$th row lists the pairs $(j, i)$ with $j \in \{1, \LL, n\}$ ranging
clockwise, starting at $i$.

Take a pair $(i,j)$. If $i = j$, then $F_i J$ is true by assumption.
If $i \neq j$, then consider the pair that precedes it in the above ordering.
It is then of the form $(i_1, j)$, where $i = i_1 \oplus 1$. By the induction
hypothesis $F_{i_1} J$ is true, so by the exit condition
$\neg K_{i_1} \neg F_{i} J$ is true.

Suppose now towards a contradiction that $\neg F_{i_1 \oplus 1} J$ is
true.  Then $i_1 \oplus 1 \neq j$.  Hence by virtue of the considered
communication mode and Definition \ref{def:model} it follows that
agent $i_i$ knows that $\neg F_{i_1 \oplus 1} J$ is true since the
only way for $i_1 \oplus 1$ to become familiar with $J$ is by means of
a call from $i_1$.  So $K_{i_1} \neg F_{i} J$ is true. This yields a
contradiction. Hence $F_{i} J$ is true.

So we showed, as desired, that $\bigwedge_{i=1}^{n}
\bigwedge_{j=1}^{n} F_i J$ is true in the considered leaf.
\end{proof}

\begin{proof}[of Proposition \ref{proposition:R2}]
  To start with, $\bigwedge_{i=1}^{n} F_i I$ is true in every node of
  the computation tree.  Suppose the exit condition
  $\bigwedge_{i=1}^{n} K_i F_{i \oplus 1} I \ominus 1 $ is true at a
  node of the computation tree (in short, true). It implies that
  $\bigwedge_{i=1}^{n} F_{i \oplus 1} I \ominus 1$ is true. Fix $i \in
  \{1, \LL, n\}$. By the above $F_i I \ominus 2$ is true.  Further,
  the implication $K_i F_{i \oplus 1} I \ominus 1 \to F_i I \ominus 1$
  is true in every node of the computation tree (remember, the agents
  are positioned on a directed ring). If $n = 3$, this proves that
  $\bigwedge_{j=1}^{n} F_i J$ is true.

\NI
If $n = 4$, we note that $K_iF_{i \oplus 1} I \ominus 1$ implies that
agent $i \oplus 1$ learned $I \ominus 1$ through a call of agent $i$
and hence the implication $K_iF_{i \oplus 1} I \ominus 1 \to F_i I
\oplus 1$ is true in every node of the computation tree, as well
(remember that the mode is push-pull). We conclude that
$\bigwedge_{j=1}^{n} F_i J$ is true.
\end{proof}

\begin{proof}[of Proposition \ref{proposition:R3}]

First, note that the following three statements are equivalent for each
node $\CSequence$ of an arbitrary computation $\xi$ and each agent $i$:
\begin{itemize}
\item $i$ is disabled at $\CSequence$,
\item $(\Model_{DR}, \CSequence) \models (\bigwedge_{j = 1}^{n} F_i J) \wedge K_i F_{i \oplus 1} I \ominus 1$,
\item a sequence of calls $(i \oplus 2, i \oplus 3), \ (i \oplus 3, i \oplus 4), \LL, (i, i \oplus1)$ (possibly interspersed with other calls) has taken place in $\xi$
before $\CSequence$.
\end{itemize}
Suppose now towards a contradiction that an infinite fair computation $\xi$
exists. We proceed by case distinction.

\noindent
\fbox{Case 1} Some agent becomes disabled in $\xi$.  

\noindent
We claim that if an agent $i$ becomes disabled in $\xi$, then also
agent $i \oplus 1$ becomes disabled in $\xi$. Indeed, otherwise by fairness
at some point in $\xi$ after which $i$ becomes disabled,
agent $i \oplus 1$ calls his successor, $i \oplus 2$, and by the above
sequence of equivalences in turn becomes disabled.

We conclude by induction that at some point in $\xi$ all agents become disabled and
hence $\xi$ terminates, which yields a contradiction.  

\noindent
\fbox{Case 2} No agent becomes disabled in $\xi$. 

\noindent
By fairness each agent calls in $\xi$ infinitely often his successor.
So for every agent $i$ there exists in $\xi$ the sequence of calls $(i
\oplus 2, i \oplus 3), \ (i \oplus 3, i \oplus 4), \LL, (i, i
\oplus1)$ (possibly interspersed with other calls).  By the above
sequence of equivalences after this sequence of calls agent $i$
becomes disabled, which yields a contradiction.
%
%
%
%
%
%
%
\end{proof}

\begin{proof}[of Proposition \ref{proposition:R4}]
\mbox{}

\NI
\fbox{Termination}
It suffices to note that after each call $(i, i
\oplus 1)$ the size of the set
\[
\{(i,j)\in \Agents \times \Agents \mid \neg K_i F_{i \oplus 1} J \}
\]
decreases.

\NI
\fbox{Correctness}
Consider a leaf of the computation tree. Then the exit condition
\[
\bigwedge_{i=1}^{n} \bigwedge_{j=1}^{n} (\neg F_i J \vee K_i F_{i \oplus 1} J)
\]
is true. As in the case of protocol R1 
we prove that it implies each $F_iJ$ is true
by induction on the pairs $(i,j)$, where $i, j \in \{1, \LL, n\}$, ordered as follows:
\begin{align*}
& (1,1), (2,1), \LL, (n,1), \\ 
& (2,2), (3,2), \LL, (1,2), \\
& \LL, \\
& (n,n), (1,n), \LL, (n-1,n).
\end{align*}
Take a pair $(i,j)$. If $i = j$, then $F_i J$ is true by assumption.
If $i \neq j$, then consider the pair that precedes it in the above
ordering, so $(i_1, j)$, where $i = i_1 \oplus 1$. By the induction
hypothesis $F_{i_1} J$ is true, so by the exit condition $K_{i_1}
F_{i} J$ is true and hence $F_{i} J$ is true.
\end{proof}

\begin{proof}[of Proposition \ref{proposition:R4f}]
Consider the following sequence of statements:
\begin{enumerate}[(i)]
\item $i$ is disabled at $\CSequence$,
\item $(\Model_{DR}, \CSequence) \models \bigwedge_{j=1}^{n} (F_i J \to K_i F_{i \oplus 1} J)$,
\item $(\Model_{DR}, \CSequence) \models  K_i F_{i \oplus 1}$,
\item a sequence of calls $i \ominus 1 \pull i, i \ominus 2 \pull i \ominus 1, \LL, i \pull i \oplus 1$ (possibly interspersed with other calls) has taken place in $\xi$ before $\CSequence$.
\end{enumerate}
It is easy to verify that these statements are logically related in the following way:
\[
\mbox{(i)} \Leftrightarrow \mbox{(ii)} \Ra \mbox{(iii)} \Ra \mbox{(iv)} \Ra \mbox{(ii)} 
\]
for each node $\CSequence$ of an arbitrary computation $\xi$ and each agent $i$.
They are therefore equivalent.
Suppose now towards a contradiction that an infinite fair computation $\xi$ exists.
As in the proof of Proposition \ref{proposition:R3} we proceed by case distinction.

\noindent
\fbox{Case 1} Some agent becomes disabled in $\xi$.  

\noindent
We claim that if an agent $i$ becomes disabled in $\xi$, then also
$i \ominus 1$ becomes disabled in $\xi$. Indeed, otherwise by fairness
at some point in $\xi$ after which $j$ becomes disabled,
agent $i \ominus 1$ calls his successor, $i$, and by the above sequence of equivalences
in turn becomes disabled.

We conclude by induction that at some point in $\xi$ all agents become disabled and
hence $\xi$ terminates, which yields a contradiction.  

\noindent
\fbox{Case 2} No agent becomes disabled in $\xi$. 

\noindent
By fairness each agent calls in $\xi$ infinitely often his successor.
So for every agent $i$ there exists in $\xi$ a sequence of calls $i
\ominus 1 \pull i, i \ominus 2 \pull i \ominus 1, \LL, i \pull i
\oplus 1$ (possibly interspersed with other calls).  By the above
sequence of equivalences, after this sequence of calls agent $i$ becomes
disabled, which yields a contradiction.
\end{proof}

\end{document}